\documentclass[10pt,conference]{IEEEtran}
\IEEEoverridecommandlockouts
\usepackage{enumitem}
\usepackage{footnote}
\usepackage{cite}
\usepackage{amsmath,amssymb,amsfonts}
\usepackage{array}

\usepackage{lscape}
\usepackage{booktabs}
\usepackage{multirow}

\usepackage{array,multirow,etoolbox,graphicx,amsmath,subcaption}
\usepackage{framed}
\usepackage[hyphens]{url}
\usepackage{listings}
\usepackage{tcolorbox}
\usepackage[normalem]{ulem}
\usepackage{authblk}
\usepackage{blindtext}
\usepackage{caption} 
\usepackage{hyperref}
\usepackage{algorithmic}
\usepackage{graphicx}
\usepackage{textcomp}
\usepackage{xcolor}
\usepackage{booktabs}
\usepackage{multirow}
\usepackage{authblk}
\usepackage{makecell}
\usepackage{tabularx}
\usepackage{caption}
\usepackage{subcaption}
\usepackage{boxedminipage}
\usepackage{graphics}
\def\BibTeX{{\rm B\kern-.05em{\sc i\kern-.025em b}\kern-.08em
    T\kern-.1667em\lower.7ex\hbox{E}\kern-.125emX}}

\newcommand{\edit}{{\tt edit2vec}}
\newcommand{\lstm}{{\tt LSTM}}
\newcommand{\bow}{{\tt Bag-of-words}}
\newcommand{\cts}{{\tt code2seq}}
\newcommand{\ctv}{{\tt code2vec}}

\title{Assessing the Effectiveness of Syntactic Structure to Learn Code Edit Representations}

\author{Syed Arbaaz Qureshi\\
  Microsoft Research India \\
  \texttt{t-syqure@microsoft.com} \\\and
  Sonu Mehta \\
  Microsoft Research India \\
  \texttt{Sonu.Mehta@microsoft.com} \\
  
  Ranjita Bhagwan \\
  Microsoft Research India \\
  \texttt{bhagwan@microsoft.com} \\\and
  Rahul Kumar \\
  Microsoft Research India \\
  \texttt{rahulku@microsoft.com}\\
  }

\begin{document}
\maketitle

\begin{abstract}
In recent times, it has been shown that one can use {\em code as data} to aid various applications such as automatic commit message generation, automatic generation of pull request descriptions and automatic program repair. Take for instance the problem of commit message generation. Treating source code as a sequence of tokens, state of the art techniques generate commit messages using neural machine translation models. However, they tend to ignore the syntactic structure of programming languages.

Previous work, i.e., \cts{} \cite{alon2018code2seq} has used structural information from Abstract Syntax Tree (AST) to represent source code and they use it to automatically generate method names. In this paper, we elaborate upon this state of the art approach and modify it to represent source code edits. We determine the effect of using such syntactic structure for the problem of classifying code edits. Inspired by the \cts{} approach,  we evaluate how using structural information from AST, i.e., paths between AST leaf nodes can help with the task of {\em code edit classification} on two datasets of fine-grained syntactic edits.

Our experiments shows that attempts of adding syntactic structure does not result in any improvements over less sophisticated methods. The results suggest that techniques such as \cts{}, while promising, have a long way to go before they can be generically applied to learning code edit representations. We hope that these results will benefit other researchers and inspire them to work further on this problem.
\end{abstract}

\begin{IEEEkeywords}
Neural network, Classification, Abstract Syntax Tree
\end{IEEEkeywords}

\section{Introduction}
\label{sec:introduction}

The recent explosion of open-source and the ready availability of online version-control systems such as GitHub~\cite{GitHubOrganization} have led to an enormous amount of code in the public domain. This in turn has given rise to the phrase, \textit{code as data}, which alludes to using code as input to machine learning models and creating tools that extract useful information from code. Deep learning methods, which have been very successful in solving problems related to natural language processing, translation, image classification and speech recognition, have recently been applied to code as well.

Take for instance the problem of creating \textit{code embeddings}. Recent work has explored how code embeddings, akin to word embeddings in the field of natural language processing (NLP)~\cite{pennington2014glove, bakarov2018survey, mikolov2013distributed}, can be used to summarize code~\cite{allamanis2016convolutional, iyer-etal-2016-summarizing}, generate documentation~\cite{leclair2019neural, movshovitz2013natural} and detect code clones~\cite{tufano2018deep, white2016deep}. Even though in theory, software can be very complex, it appears that even a small statistical model can capture regularity in natural software to a great extent (naturalness hypothesis)~\cite{allamanis2018survey}. These findings have encouraged the use of embedding-based techniques on code.

\begin{figure*}
\includegraphics[width=\textwidth]{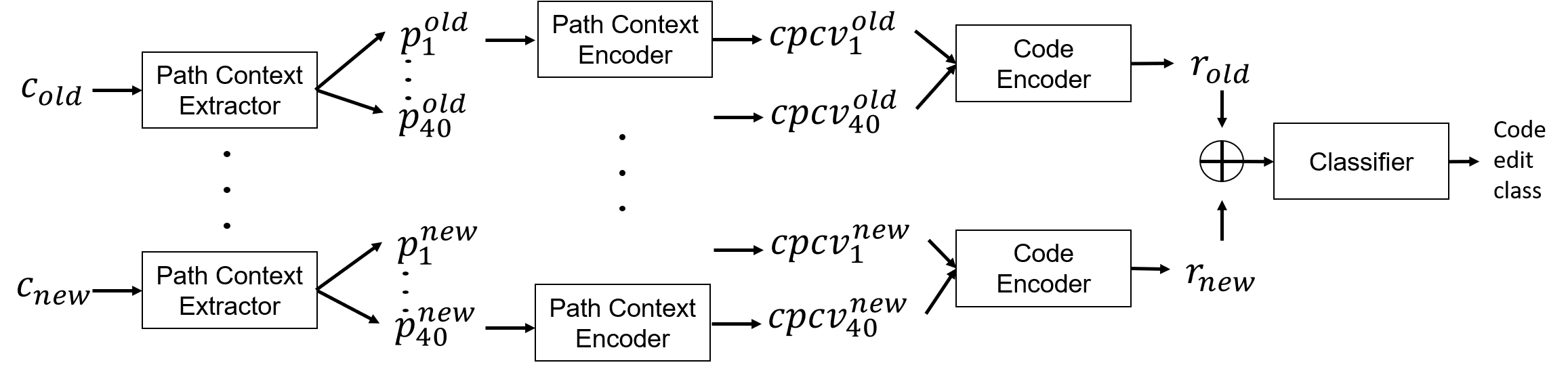}
\caption{\edit{} model architecture}
\label{fig:e2vdesign}
\end{figure*}

However, while the results of these recent efforts have been encouraging, we recognize that there are significant differences between natural language and code. Along with syntactic structure, code statements have long range correlations, much larger vocabulary than natural language and reduced robustness to minor changes~\cite{allamanis2018survey}. Previous work attempts to capture such long range correlations by using \textit{syntactic structure}: data and control-flow graphs of code \cite{allamanis2017smartpaste, park2012using, nobre2016graph}, abstract syntax trees~\cite{zhang2019novel, allamanis2017learning} etc. It has largely focused on applying such techniques on whole code snippets (code snapshots) like individual methods or classes~\cite{alon2019code2vec, alon2018code2seq}.
 
In this paper, we ask the question, ``\textit{Are techniques that capture syntactic structure useful in understanding code edits, rather than code snippets?}". Some fundamental differences between code edits and code snippets should be brought to light. Code edits do not have a fixed context; they can span multiple files and different methods. On the other hand code snippets have a well-defined context, especially as explored in previous work, where they span only a single method of class~\cite{alon2019code2vec, alon2018code2seq}. 
%Also, the objective of a code edit can be described at a high-level as, for instance, adding a feature or fixing a bug, while a code snippet can have much more specific summary statements (such as `save text to a file'). 
 
Our work is inspired by \cts{}, which leverages the syntactic structure of programming languages to summarize a given code snippet. It is language agnostic. It represents a given code snippet as a set of path-contexts over its abstract syntax tree (AST), where each path is compressed to a fixed length vector using bi-directional LSTMs. It then uses the attention mechanism to get a weighted average of the path vectors to represent the code snippet in a vector space, and uses it to produce the code summary or caption.

In this paper, we use an approach similar to \cts{} to learn distributed representations of code edits and use it for fine-grained code edit classification. We concentrate on two specific tasks: bug-fix classification (Section \ref{bugfixclassification}) and code transformation classification (Section \ref{ctclassification}). \cts{} is reported to be the state of the art on the task of method name generation~\cite{ alon2018code2seq}, and the authors of \cts{} claim that it is generalizable. This motivates us to test \cts{} on code edit classification. We compare our approach to represent code edits with two other models: a) considering code as a collection of tokens in a \bow{} model and b) considering code as a sequence of tokens using an \lstm{} model.

Our results suggest that using syntactic structure such as path contexts \textbf{\textit{does not}} help improve the efficacy of code edit classification. Based on our experience, we observe that:

\noindent
a) Previous-work using syntactic structure~\cite{alon2018code2seq, alon2019code2vec} concentrated on method name prediction, and therefore benefited heavily from the identifier names which are captured as part of the terminal nodes in AST. Code edit classification, on the other hand, does not benefit from specific identifier names. 

\noindent
b) Training models for classifying code edits with syntactic constructs may require significantly more data than is currently available: \cts{} used more than $15$ million data-points for training while we have access to edit datasets of size in the order of tens of thousands. We believe that it is fundamentally difficult to collect millions of labelled code edits for reasons we describe in section~\ref{sec:discussion}.

We therefore conclude that using syntactic structure, while effective on code snippets, has a long way to go before it can be applied to learning code edit representations.

The main contributions of this paper can be summarized as follows:
\begin{itemize}
    \item We propose a new code edit classification approach that uses syntactic structure of code along with the attention mechanism to learn distributed representation of code edits. We use this to classify bug-fixes and code transformations.
   
    \item We create a new code edit dataset generated by a set of C\# code transformers named Roslynator analyzers\cite{RoslynatorAnalyzers} from the top $250$ C\# GitHub repositories based on their popularity and release the data at \url{https://figshare.com/s/e2e4c6762c696825c6d1}.
    
    \item We conduct experimental evaluations using two tasks: bug-fix classification on a data-set of Java bug-fixes (ManySStuBs4J dataset~\cite{karampatsis2019often}), and code transformation classification using data obtained from the top $250$ C\# GitHub repositories based on their popularity. Our results provide promising empirical evidence that the baseline approach using LSTMs outperforms the approach that uses syntactic structure of code.
\end{itemize}

%XXXXXX We can get rid of the below para

%The rest of the paper is organized as follows: Section \ref{motivation} explains why code edit classification is an important task. Section \ref{sec:network} describes the details of different model architectures used for the code edit classification task. Section \ref{sec:exp} provides an overview of the two code edit classification tasks and the datasets used in our analysis. Section \ref{sec:evaluation} describes evaluation on the two datasets. In Section \ref{sec:discussion}, we discuss the threats to validity of our approach. Section \ref{sec:relatedwork}  gives an overview of the related work in this space and finally, in Section \ref{sec:conclusion}, we conclude with a call for further research in the area of code edit classification.
\section{Motivation}
\label{motivation}

%The usage and contribution to open-source software (OSS) has increased considerably over the past years. Developers from across the world create thousands of commits/code edits/bug-fixes everyday. 

Code edits fall into various categories such as bug-fixes, feature additions,  and code refactors. With bug-fixes, there is fine-grained classification such as those based on the bug-template, as defined in Section \ref{bugfixclassification}. Learning representations of such code edits is useful for many tasks such as bug-fix classification (classifying which bug-fix template is applied on the code), commit message generation and recommendation systems for automatic program repair. However, class labelled data for code edits is relatively scarce, especially for big edits spanning over multiple lines. Recently, researchers have created datasets like ManySStuBs4J for bug-fixes in Java \cite{karampatsis2019often}. This has enabled us to concentrate on fine-grained bug-fix classification in this paper. Here is an example of one such fine-grained bug-fix. Consider a scenario, where we have two classes, {\tt class1} and {\tt class2}, that inherit from the same parent class. Both these classes override a method from the parent class, and have their own implementation of this method. A developer has mistakenly called this method from {\tt class1}, when it should be called from {\tt class2}. The developer does not notice this error at compile time, but she later realizes her mistake and fixes this bug. Previous work has shown that such bug-fixes, which modify a small set of lines of code, are seen quite frequently during development phase~\cite{karampatsis2019often}.

Automatically classified bug-fixes can help the process of software development in many ways. For instance, we can use this information to decide the right set of reviewers to review a specific kind of bug-fix. It can be used for test selection; only tests relevant to the specific bug-fix need to be run. Also, it can be used for vulnerability management tasks, which is one of the most urgent security challenges faced by software community. As defined in \cite{lozoya2019commit2vec}, vulnerability management tasks ensure that the open-source components included in various products of a company are free from (known) vulnerabilities. Currently, the National Vulnerability Database is used to keep track of disclosed vulnerabilities, but it has consistency and coverage issues. Moreover, this manual approach cannot scale with the increase in open-source code.

%XXXXX Why are we even speaking about commit message generation? We can save a lot of space if we get rid of commit message generation.

%Commit messages, which summarize code edits, are essential for program comprehension and maintainability. In general, developers find it difficult to write meaningful commit messages either due to lack of time or direct motivation \cite{dyer2013boa}. Learning code edit representations can be helpful in generating meaningful commit messages automatically; we can use machine translation techniques over the learned code edit representations, to generate the message.

Automatic program repair \cite{campbell2014syntax, bhatia2016automated} is another use-case of learning code edit representations, where they are used to automatically generate code-patches or bug-fixes to solve issues similar to those which have occurred in the past \cite{yin2018learning}.

%XXX Can shorten the above two paraXXX

%Broadly these tasks can be categorized into two categories  1) \textit{classification} and 2) \textit{generation}. In general, classification tasks are simplified versions of generation tasks. Generation can be considered as classification with extremely large number of classes. Models that seem to perform not so good in classification tasks perform worse in generation tasks.

We conduct experiments to evaluate our proposed model on simple classification tasks which have a deterministic solution. The aim of our work is not to solve the problem of bug-fix classification, but to evaluate if code embeddings learnt using syntactic structure of code help improve over baseline models that consider code as natural language, ignoring it's syntactic structure.

% Jacob Devlin, Jonathan Uesato, Rishabh Singh, and Pushmeet Kohli. Semantic code repair using neuro-symbolic transformation network
%Neural program repair by jointly learning to localize and repair
% L. Gazzola, D. Micucci, and L. Mariani. Automatic software repair: A survey. IEEE Transactions on Software Engineering, pp. 1–1, 2018
% C. Le Goues, N. Holtschulte, E. K. Smith, Y. Brun, P. Devanbu, S. Forrest, and W. Weimer. The manybugs and introclass benchmarks for automated repair of c programs. IE

% Claire Le Goues, ThanhVu Nguyen, Stephanie Forrest, and Westley Weimer. GenProg: A Generic  Method for Automatic Software Repair. IEEE Trans. Software Eng., 38(1):54–72, 2012.
% Martin Monperrus. Automatic software repair: A bibliography. ACM Comput. Surv., 51(1):17:1–% 17:24, 2018.
% Manish Motwani, Sandhya Sankaranarayanan, Rene Just, and Yuriy Brun. Do automated program  repair techniques repair hard and important bugs? Empirical Software Engineering, 23(5):2901–% 2947, 2018

%XXX if we can put references that would be good, also in the above para for vulnerability management etc. XXX

\section{Model Architecture}
\label{sec:network}
% \begin{table*}[t]
%  \centering
%  \begin{tabular}{|c|c|}
%  \hline
%  Notation & Definition \\ 
%  \hline\hline
%  $c_{old}$/$c_{new}$ & source code (sequence of code  \\ & tokens) before/after the edit \\
%  \hline
 
%  $t_{old}$/  $t_{new}$ & abstract syntax tree(AST) for $c_{old}$/$c_{old}$ \\
%  \hline
%  $P_{old}$/ $P_{new}$  & set of path-contexts obtained from $t_{old}$/$t_{new}$ \\
%   \hline
% %  $P_{new}$ & set of path-contexts obtained from $t_{new}$ \\
% %  \hline
%  $p^{old}_{i}$/$p^{new}_{i}$ & one of the path-contexts from the set $P_{old}$/$P_{new}$  \\ 
%  \hline
%  $CPCV^{old}_i$/$CPCV^{new}_i$ & Compact Path Context Vector encoding a path context $p^{old}_{i}$/$p^{new}_{i}$ 
%  \\ 
%  \hline
% $r_{old}$ /$r_{new}$ & 160-dimensional vector representing $c_{old}$ /$c_{old}$ using \edit{} model \\
% \hline
% $r^{LSTM}_{old}$ /$r^{LSTM}_{new}$ & 144-dimensional vector representing $c_{old}/$ /$c_{new}$ using \lstm{} model  \\
%  %\hline\hline
%  %code2seq-based model (canonicalized) & 98.44\% \\ 
%  %\hline
%  %LSTM model (canonicalized) & 99.21\% \\
%  %\hline
%  %LSTM-attention model (canonicalized) & 99.17\% \\
%  \hline
%  \end{tabular}
%  \vspace{0.1in}
%  \caption{List of all the symbols used}
% \label{tab:notation}
% \end{table*}

\begin{table*}[t]
 \centering
 \begin{tabular}{|c|c|}
 \hline
 Old code & New code\\ \hline
 $c_{old}$ = \texttt{processURL(message, depth, \textit{baseURL}, \textbf{url})} & $c_{new}$ = \texttt{processURL(message, depth, \textbf{url}, \textit{baseURL});}\\

 \hline
 $t_{old}$ = \includegraphics[scale = 0.35]{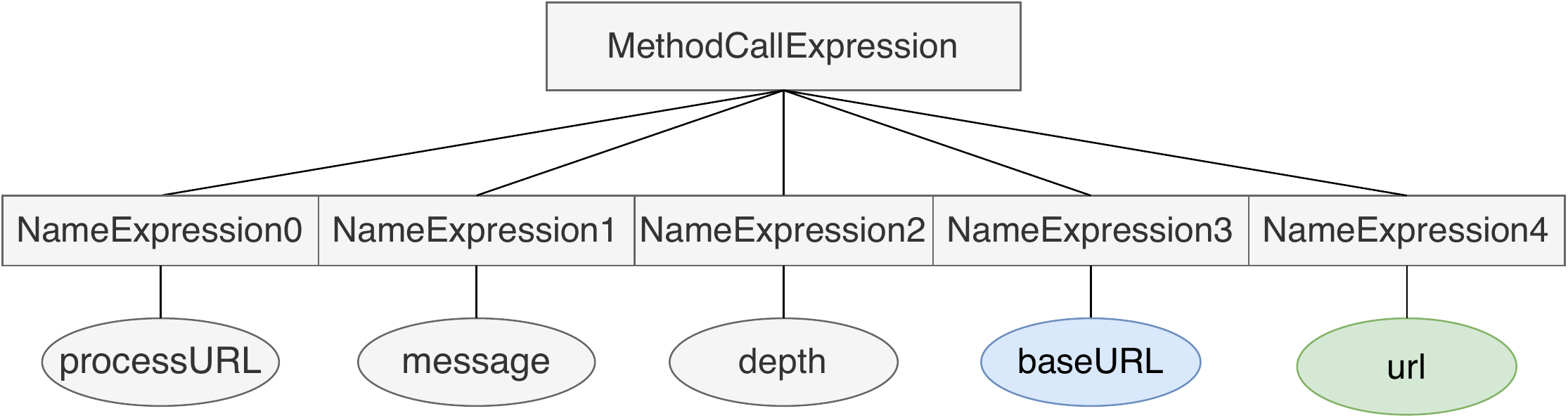}
 & $t_{new}$ = \includegraphics[scale = 0.35]{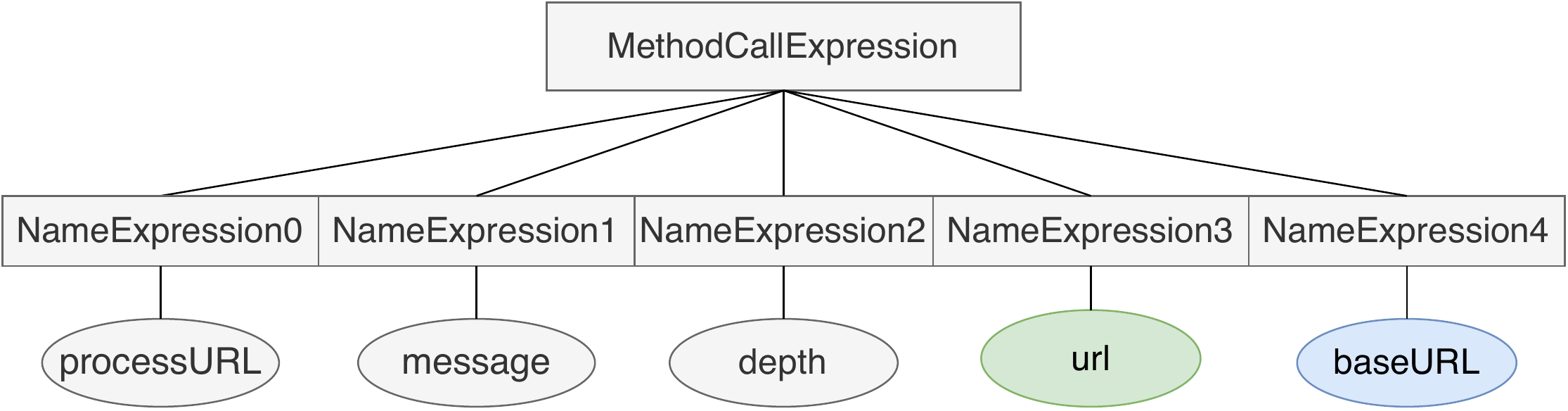} \\
 \hline
 $P^{old}$ =\{ \texttt{{\em processURL}, NE0, MCE, NE3, {\em baseURL}} &  $P^{new}$ =\{ \texttt{{\em processURL}, NE0, MCE, NE4, {\em baseURL}} \\
 \texttt{{\em processURL}, NE0, MCE, NE2, {\em depth}} & \texttt{{\em processURL}, NE0, MCE, NE2, {\em depth}} \\
 \texttt{{\em message}, NE1, MCE, NE2, {\em depth}} ...\}& \texttt{{\em message}, NE1, MCE, NE2, {\em depth}} ...\}
 \\
 \hline
PCE Output  = $\{CPCV^{old}_1$, $CPCV^{old}_3$, .... $CPCV^{old}_{40}\}$.  & PCE Output  = $\{CPCV^{new}_1$, $CPCV^{new}_3$, .... $CPCV^{new}_{40}\}$  \\
 \hline
CE output = $r_{old}$ [$160$-D vector] & CE output = $r_{new}$ [$160$-D vector]\\
\hline
 \end{tabular}
 \vspace{0.1in}
 \caption{Example to explain Path Context Encoder. The names of the non-leaf tokens are abbreviated; for example, \texttt{NE0} stands for \texttt{NameExpression0}.}
\label{tab:pathvecex}
\end{table*}

We have built three models to evaluate whether syntactic structure captured using AST helps learn better representations of code edits. All of these models are classifiers, and they differ in the way we represent the input set of old and new code snippet. In this section, we describe the details of these three models and their construction.

% \begin{itemize}
%     \item {\em \edit:} This model uses the syntactic structure of code by representing it as a set of paths between different terminal nodes of AST.
%     \item {\em \lstm:} This model considers the code as a sequence of tokens and ignores its syntactic structure. 
%     \item {\em \bow:} This model uses a \bow{} approach and ignores the sequence of the tokens present in a code snippet, and its syntactic structure.
% \end{itemize}

%We compare the code2seq architecture with two other architectures - LSTM and LSTM with attention. We describe the three architectures as follows.

\subsubsection{\edit{}}

This model uses the syntactic structure of code by representing it as a set of paths between different terminal nodes of AST, similar to the way syntax is captured in \cts{} model. The \cts{} model \cite{alon2018code2seq} was designed to generate method names from the method body. It followed the encoder-decoder architecture for Neural Machine Translation (NMT). While we draw inspiration from this work, our model differs significantly from \cts{} in two ways:

\begin{itemize}
    \item{\em Characterizing the code edit.} The input to the model has to capture the difference between the code before the edit and after the edit. The \cts{} model did not need to do this since the input was just a code snippet containing the method body.
    \item{\em Classification, not generation.} Our task is one of classification, not generation. Hence we replace the decoder layer of \cts{} with a classifier. In our implementation, we have used a \texttt{softmax} layer for multi-class classification.
\end{itemize}

Figure~\ref{fig:e2vdesign} shows the overall design of the \texttt{edit2vec} model. It should be noted that this model is language agnostic, and can be used with code changes of any length in a single file. We first explain how we represent a code edit as an input to our model.

\paragraph{\textbf{Path-Context Extractor}}
\label{path-context}
We characterize a code edit as the pair $\{c_{old}, c_{new}\}$ where $c_{old}$ is the source code before the edit and $c_{new}$ is the source code after the edit. Corresponding to $\{c_{old}, c_{new}\}$, we build the Abstract Syntax Tree (AST) denoted by $\{t_{old}$ and $t_{new}\}$ respectively. Similar to \cts{}, we use the {\em path-context} as a means to capture the syntactic structure of code. For a given AST, the path-context is the shortest path from one terminal node (leaf) to another. We represent a path-context as sequences of terminal and non-terminal nodes; it starts with one terminal node, which is also called the {\em left-context}, then a set of non-terminal nodes, which we call as the {\em path}, follow. The last node, or the {\em right-context}, is the other terminal node of the path.

The source code can represent the method, the class or the file which was edited. An AST for such a source code can be large and hence it may contain a large number of path-contexts. Randomly selecting a certain number of path-contexts (as done in \cts{}) might not capture the edit completely. Since we are dealing with smaller code edits, typically $1$-$2$ lines of code, we restrict our source code to the lines of code that are changed. In addition to that, we filter out data-points which have more than $40$ path-contexts in $t_{old}$ or $t_{new}$. We have decided to select $40$ path-contexts as majority of the examples in our dataset have less than $40$ path-contexts. In case the AST is very small and has fewer than $40$ path-contexts, we pad the path-contexts with dummy values, to make the input size uniform \footnote{\url{https://www.tensorflow.org/guide/keras/masking_and_padding}}. Note that, this model can be used with code changes of any length in a single file. However, we cannot use this model for changes spanning over multiple files.

In this way, for each pair of ASTs ${t_{old}, t_{new}}$, we get ${P_{old}, P_{new}}$, where $P_{old} = \{p^{old}_1, \ldots, p^{old}_{40}\}$ is the set of $40$ path-contexts from the AST $t_{old}$ and $P_{new} = \{p^{new}_1, \ldots, p^{new}_{40}\}$ is the set of $40$ path-contexts from the AST $t_{new}$. For example, there is a code edit where the arguments of a method are swapped. Table \ref{tab:pathvecex} shows how to get $\{P_{old},  P_{new}\}$ from $\{c_{old}, c_{new}\}$ for this example.

%Each path-context is converted into a $d$-dimensional vector by the first component in the $\edit{}$ model, which we call the {\em Path-Context Encoder}.

%We use an attention mechanism to combine $P_{old}$ and $P_{new}$  into one vector using Code Encoder~\cite{XXX} XXX what is Code Encoder XXX which outputs a pair of vectors  $\{v_{old}, v_{new}\}$. Finally these two vectors are concatenated and passed to a classifier as input, the details of which are described in Section~\ref{XXX}. XXX do we associate two path-contexts with each other? This is not clear XXX.

%We now describe the key components of the \syntax model. Broadly the model can be divided into the following three components:
%\begin{itemize}
 % \item Path-Context Encoder
 % \item Code Encoder
  %\item Classifier
%\end{itemize}

\paragraph{\textbf{Path-Context Encoder (PCE)}}
%https://www.overleaf.com/project/5e6f55d19f8ad70001cf9a7b
%We haven't described what a path-context is.

\begin{figure}
\includegraphics[scale = 0.7]{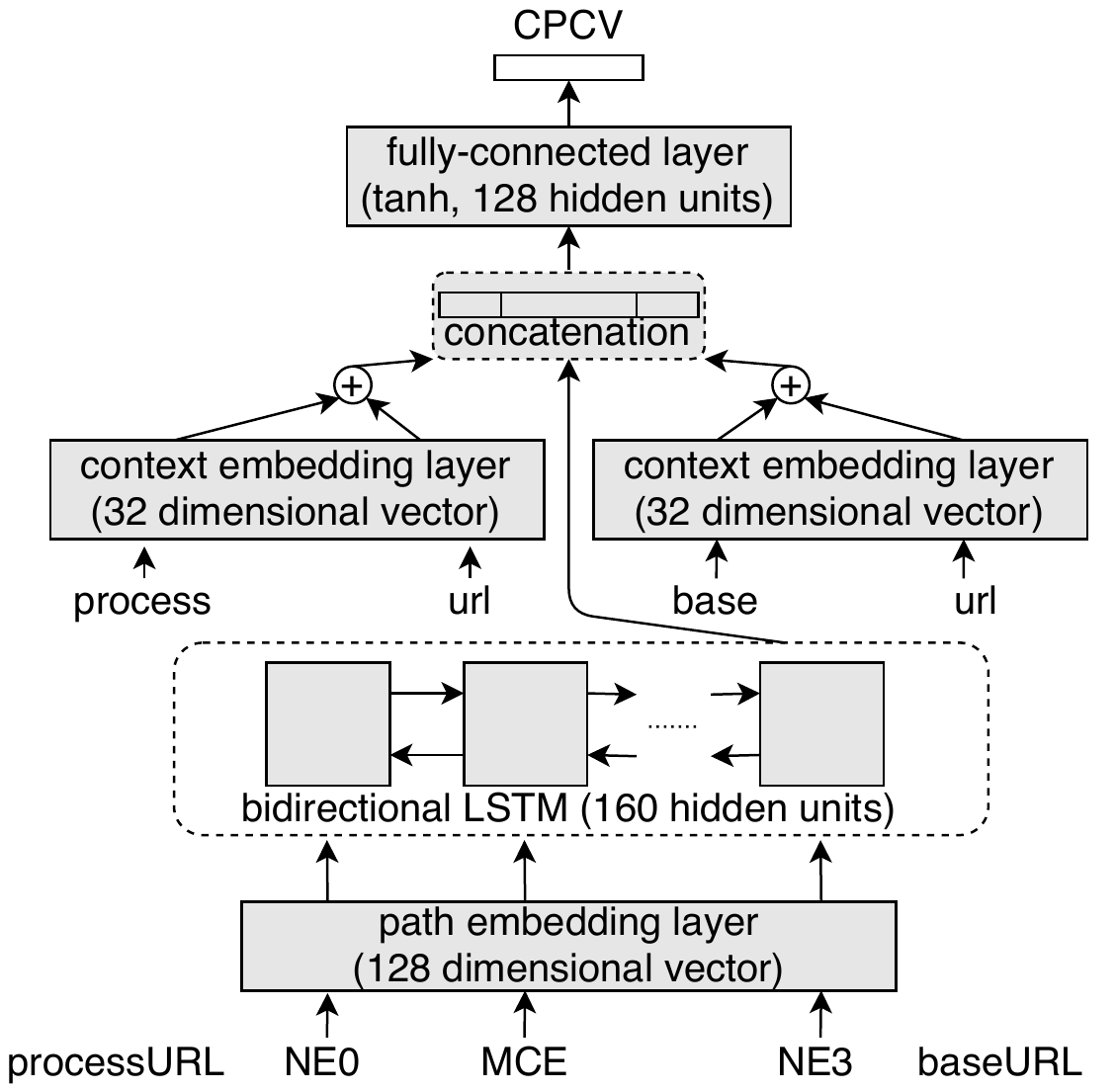}
\caption{Path Context Encoder}
\label{fig:path_context_encoder}
\end{figure}

These path-contexts are input to the {\em Path-Context Encoder (PCE)}. This uses a similar technique as \cts{} to encode a path-context into a $128$-dimensional vector called $CPCV$ (Compact Path-Context Vector). Figure \ref{fig:path_context_encoder} shows the model architecture of the Path-Context Encoder. Consider a path-context $p$ of Table \ref{tab:pathvecex}: \texttt{processURL}, \texttt{NE0}, \texttt{MCE}, \texttt{NE3}, \texttt{baseURL}. The PCE encodes the path-context in the following way:

For terminal nodes (\texttt{processURL} and \texttt{baseURL}) in the path-context, i.e., the left and right contexts, the encoder first divides them into {\em sub-tokens}; the token `{\tt processURL}' splits into `{\tt process}' and `{\tt url}'. The PCE uses a $32$-dimensional embedding vector to represent each sub-token, and then averages the embedding vectors of the sub-tokens to obtain the final vector for this terminal node. This generates two vectors $v_{processURL}$ and $v_{baseURL}$.

For non-terminal nodes (\texttt{NE0}, \texttt{MCE}, \texttt{NE3}), the PCE first embeds them to three $128$-dimensional embedding vectors, and feeds them to a bi-directional LSTM layer \cite{10.1162/neco.1997.9.8.1735} of $160$ hidden units. The output of this LSTM layer, which we call $v_{path}$, thus encodes the path into a $160$-dimensional vector.

The {\em PCE} concatenates the three vectors $v_{processURL}$, $v_{path}$ and $v_{baseURL}$ and feeds this to a fully-connected \texttt{tanh} layer, which consists $128$ hidden units. We refer to the output of this layer as the {\em Compact Path-Context Vector (CPCV)}. The {\em CPCV} is a vector representation of the input path-context $p$.

\paragraph{\textbf{Code Encoder (CE)}}

\begin{figure}
\begin{center}
\includegraphics[scale=0.5]{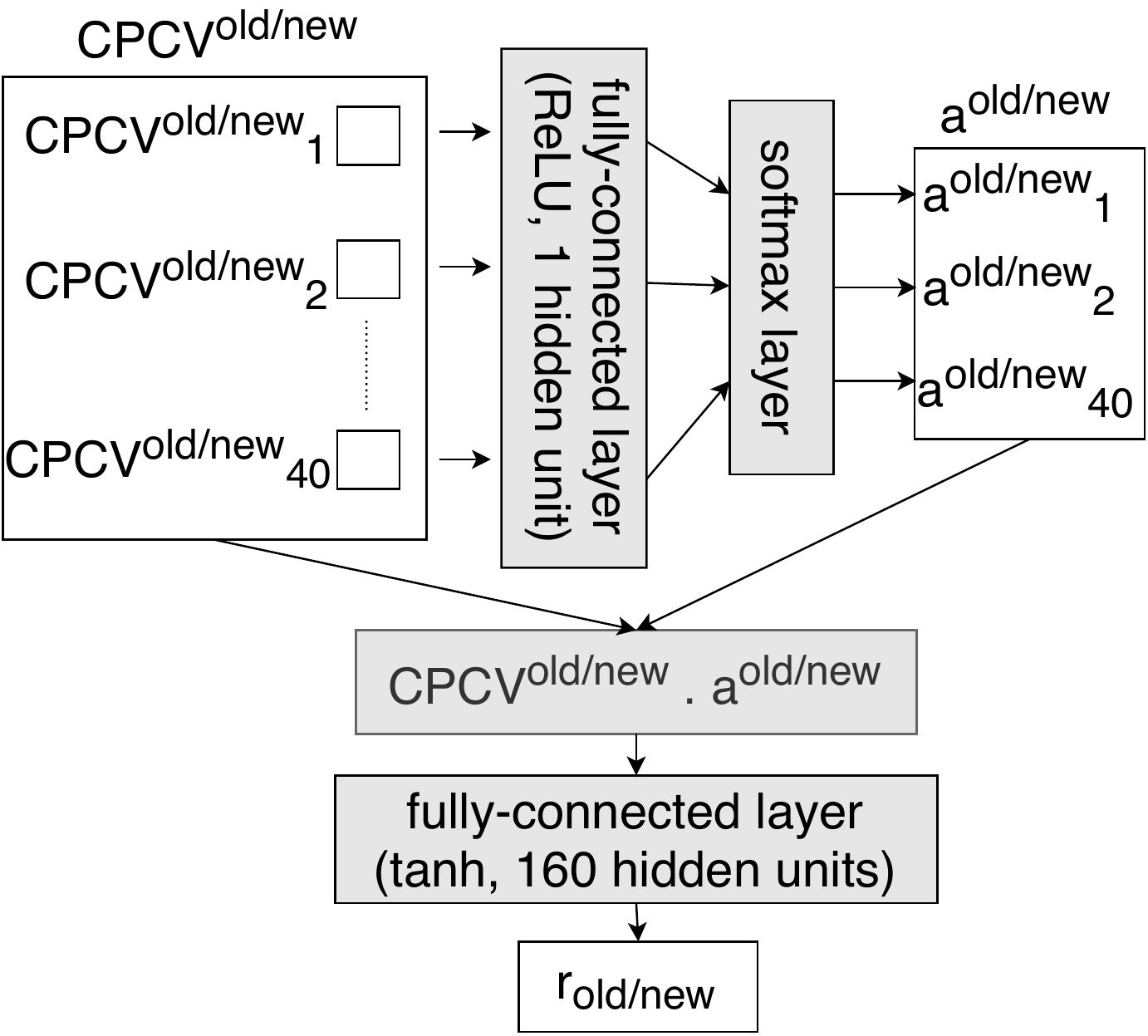}
\caption{Code Encoder}
\label{fig:code_encoder}
\end{center}
\end{figure}

Since there are not more than $40$ path-contexts each for the old and new code $\{c_{old}, c_{new}\}$, the PCE outputs $40$ CPCVs each for $c_{old}$ and $c_{new}$. Each set of $40$ CPCVs is input to {\em Code Encoder (CE)}, which uses the attention mechanism described in the \cts{} model to encode each set of CPCVs into $160$-dimensional vector. We represent these vectors as $\{r_{old}, r_{new}\}$ corresponding to $\{c_{old}, c_{new}\}$ .
Figure~\ref{fig:code_encoder} shows the detailed model architecture for Code Encoder. The input is passed to \texttt{ReLu} layer followed by \texttt{softmax} layer which outputs the attention weights ($a_1$, $a_2$, \ldots $a_{40}$) for each $CPCV$. A weighted summation of CPCVs ($CPCV_1 \times a_1$ + $CPCV_2 \times a_2$ + \ldots + $CPCV_{40} \times a_{40}$) is passed through a fully connected \texttt{tanh} layer to output 160-D vector. Code Encoder is similar to what is used in \cts{} model. Hence, in the interest of space, we do not describe it in detail. 

%Let \textbf{CPCV} be the set of CPCVs \(\{ CPCV\textsubscript{1}, CPCV\textsubscript{2}, ..., CPCV\textsubscript{40} \}\) for a code snippet. Each of the CPCVs from the set are fed to a `ReLU' fully-connected layer, which outputs a scalar value. The $40$ scalars, obtained by passing  $m$ CPCVs to this layer are normalized using a `softmax' layer. We call these normalized values as the attention values \(\{a\textsubscript{1}, a\textsubscript{2}, ..., a\textsubscript{m}\}\). We then take the weighted summation of the $m$ CPCVs, where the weights are the attention values. This weighted summation is fed to a `tanh' layer. The output of this layer acts as the encoding for the input set \textbf{CPCV}. We call this output vector as the Code Representation (r\textsuperscript{AST}).

% The two sets \textbf{CPCV}\textsubscript{old} and \textbf{CPCV}\textsubscript{new} are fed to the Code Encoder, to obtain two vectors r$^{AST}_{old}$ and r$^{AST}_{new}$. In essence, r$^{AST}_{old}$ and r$^{AST}_{new}$ are the vector representations of the old and new codes. These two vectors are used for classifying the edit, as described in the following subsection.

\paragraph{\textbf{Classifier}}
\label{sec:classfier}
The vectors $r_{old}$ and $r_{new}$, obtained from the {\em Code Encoder}, are concatenated and passed through a classifier. The classifier is a neural network with a \texttt{tanh} layer of $80$ hidden units, followed by a \texttt{softmax} layer, which outputs the class of the code edit.

\paragraph{\textbf{Model Hyperparameters}}
The entire \edit{} model is trained end-to-end for $100$ epochs, minimizing categorical cross-entropy loss using the `Adam'~\cite{kingma2014adam} optimizer. In order to reduce over-fitting we use dropout layers in the bi-directional LSTM layer, one after the \texttt{tanh} layer of the {\em Path-Context Encoder} (dropout rate = $0.2$), one after the \texttt{tanh} layer of the {\em Code-Encoder} (dropout rate = $0.4$), and one after the \texttt{tanh} layer of the Classifier (dropout rate = $0.6$). The dropout rates, the number of hidden units in each layer, the vector dimensions for path ($128$-D) and context ($32$-D) token embeddings, etc. are considered as model hyperparamaters. The values of these hyperparameters are chosen after training and evaluating $400$ \edit{} models with various combinations of hyperparameter values from a grid of values and then selecting the one that gives the best accuracy on the validation set in cross-validation~\cite{hastie2009elements}.

% The entire \edit{} model is trained end-to-end for 100 epochs, minimizing categorical cross-entropy loss using the `adam' optimizer. In order to reduce over-fitting we use dropout layers with a dropout rate of 0.2 in the bi-directional LSTM and after the `tanh' layer of the {\em Path-Context Encoder}, one after the `tanh' layer of the {\em Code-Encoder} with a dropout rate of 0.4, and another one after the `tanh' layer of the Classifier with a dropout rate of 0.6. The best set of hyperparameters (the dropout rates, the number of hidden units in each layer, the vector dimensions for path (128-D) and context (32-D) token embeddings, etc.) were chosen after training and evaluating 400 different models with various combinations of values from a grid of values.
%Once the best set of hyperparameters were identified, we perform 10-fold cross validation on the training set, where we preserve models that give best accuracy on the validation folds. More details about train-test split for the two data-sets is given in section \ref{sec:exp}.

\subsubsection{\lstm{}}
\begin{figure}
\begin{center}
\includegraphics[scale=0.6]{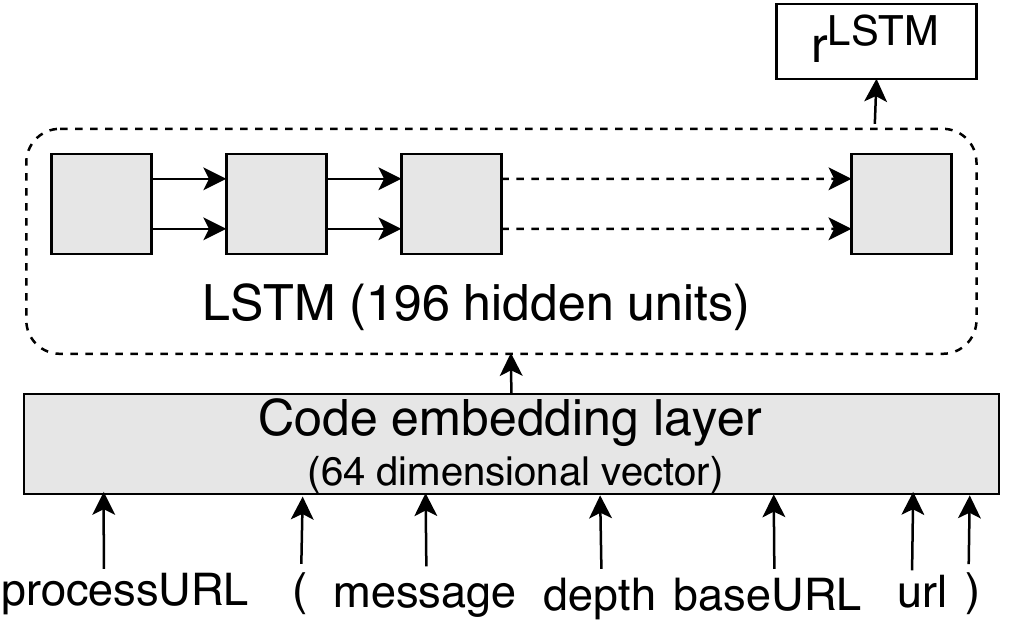}
\caption{LSTM}
\label{fig:code_LSTM}
\end{center}
\end{figure}

To compare with \edit{}, we build an LSTM-based model (\lstm{})~\cite{10.1162/neco.1997.9.8.1735} that considers code as a sequence of tokens and ignores the syntactic structure.  Given $\{c_{old}, c_{new}\}$, we tokenize each to a set of tokens. Each code token is then mapped to a $64$-dimensional embedding vector. For both $\{c_{old}, c_{new}\}$ , the sequence of tokens is input to a standard \texttt{LSTM} layer with $196$ hidden units followed by a dropout layer (dropout rate = $0.8$). Figure \ref{fig:code_LSTM} describes the architecture of the \texttt{LSTM} model. The model outputs two $196$-dimensional vectors $r^{LSTM}_{old}$ and $r^{LSTM}_{new}$ corresponding to $\{c_{old}, c_{new}\}$ respectively. The vectors $\{r^{LSTM}_{old}$, $r^{LSTM}_{new}\}$ are concatenated and then passed to a classifier. The classifier used here is same as defined in Section \ref{sec:classfier}.

\subsubsection{\bow{}}

This is the baseline model based on classical machine learning techniques. This model ignores both the syntactic structure as well as the sequence of the code tokens. It considers the code as a bag of words, regardless of their sequence in the code snippet.

Given $c_{old}$ and $c_{new}$, we tokenize each to a set of tokens. We fit two different \bow{} based vectorizers, the count-based vectorizer\footnote{\raggedright \url{https://scikit-learn.org/stable/modules/generated/sklearn.feature_extraction.text.CountVectorizer.html##sklearn.feature_extraction.text.CountVectorizer}} and the tf-idf vectorizer\footnote{\raggedright \url{https://scikit-learn.org/stable/modules/generated/sklearn.feature_extraction.text.TfidfVectorizer.html}}, on the merged set of tokens. Using the learnt vectorizer, we then separately vectorize $c_{old}$ and $c_{new}$, concatenate the two vectors and pass it to a classifier, to classify the edit. We test two different classifiers - the linear-kernel and the RBF-kernel support vector machine (SVM) classifiers\footnote{\url{https://scikit-learn.org/stable/modules/svm.html}}. The count-based vectorizer gives more weightage to the word having high frequency where as in the tf-idf vectorizer, the weightage increases with the frequency of the token but is offset by the number of code snippets in which the token appears, to adjust for the fact that some tokens appear more frequently than others (for example: datatypes in code).

\begin{table*}[t]
 \centering
 \begin{tabular}{|p{3cm}|p{8.5cm}|p{2cm}|}

% \begin{tabular}{|c|c|c|}
 \hline
 Bug-template & Description & No of samples \\ [0.5ex] 
 \hline
 \textit{change caller in function call}
 & Checks whether in a function call expression the caller object for it was replaced with another one.
 & 1488 \\ 
 \hline
 \textit{change numeral}
 &  Checks whether a numeric literal was replaced with another one & 4779 \\
 \hline
 \textit{change operand}
 & Checks whether one of the operands in a binary operation was wrong. & 741 \\
 \hline
 \textit{change operator} & Checks whether a binary operator was accidentally replaced with another one of the same type & 1711 \\
 \hline
 \textit{different method same args}
 & Checks whether the wrong function was called. & 9383 \\
 \hline
 \textit{less specific if}
 & Checks whether an extra condition which either this or the original one needs to hold ($||$ operand) was added in an if statement’s condition. & 2095 \\
 \hline
 \textit{more specific if}
 & Checks whether an extra condition ($\&\&$ operand) was added in an if statement’s condition & 1836 \\
 \hline
 \textit{overload method deleted args}
 &  Checks whether an overloaded version of the function with less arguments was called & 1040 \\
 \hline
 \textit{overload method more args}
 & Checks whether an overloaded version of the function with more arguments was called & 3820 \\
 \hline
 \textit{swap arguments}
 & Checks whether a function was called with two of its arguments swapped & 536 \\
 \hline
 \textit{swap boolean literal}
 &  Checks whether a Boolean literal was replaced & 1531 \\
 \hline
 \end{tabular}
 \vspace{0.1in}
 \caption{Descriptions of various SStUB templates, and the number of samples associated with the template.}
\label{tab:javadataset}
\end{table*}

Similar to \edit{} and \lstm{} models, we use both $c_{old}$ and $c_{new}$ as input to this model and not just the diff tokens (tokens that were either added or deleted) so as to capture the context around the code edit. Consider these two code edit examples:

Example 1: \texttt{os.file(path)} $\rightarrow$  \texttt{os.folder(path)}
Example 2: \texttt{file.getSize()}  $\rightarrow$ \texttt{folder.getSize()}

In both the cases, the set of diff tokens are \{\texttt{file}, \texttt{folder}\}, but Example 1 belongs to class `\textit{different method same args}' whereas Example 2 belongs to `\textit{change caller in function call}'. Using both $c_{old}$ and $c_{new}$ help generalize \bow{} and distinguish such code edits.

% XXXXXXXXX
% We choose to input both old and new code tokens, and not just the diff tokens (tokens that were deleted or added during the transition from old code to new code). Inputting diff tokens engulfs vital edit information, as edits of different types may have the same diff tokens. Consider the following example:

% old code 1: \texttt{os.file(path)}

% new code 1: \texttt{os.folder(path)}

% old code 2: \texttt{file.getSize()}

% new code 2: \texttt{folder.getSize()}

% In both the cases, the diff tokens are \texttt{file} and \texttt{folder}, but the edit type of the first example is `\textit{DIFFERENT METHOD SAME ARGS}' where as the edit type of the second example is `\textit{CHANGE CALLER IN FUNCTION CALL}'. This example shows that using only diff tokens engulfs vital edit information.
%XXX Talk more about tf-idf etc.

\section{Experimental Setup}
\label{sec:exp}

In this section, we describe the two code edit classification tasks along with the two datasets we have used to evaluate the three models mentioned in Section~\ref{sec:network}.

\subsection{Code Edit Classification Task}
\label{editclassificationtask}

Given the source code before edit ($c_{old}$) and source code after edit ($c_{new}$), the code edit classification task is to predict the class of the edit that was applied on $c_{old}$ to generate $c_{new}$. We analyze the following two instances of code edit classification:

\subsubsection{Bug-fix Classification}
\label{bugfixclassification}

The \textit{ManySStuBs4J data-set}~\cite{karampatsis2019often} consists of $63,923$ labelled single-line bug-fix changes mined from $1000$ popular open-source Java projects. The bug-fixes are classified into one of $16$ bug templates (the simple stupid bug templates - SStuB). Given $c_{old}$ and $c_{new}$, this task is to predict the bug-template that was applied on $c_{old}$ to generate $c_{new}$.

\paragraph{\textbf{Filtering And Pre-processing.}}
We first tokenize the source code snippets into tokens such as identifiers, keywords, constants, symbols and other elements. We use Python's {\tt javalang}\footnote{https://pypi.org/project/javalang/} module for tokenizing and parsing the code. We remove all data-points where either the $c_{old}$ or $c_{new}$ were not tokenizable by the {\tt javalang} module. This filters out $3,739$ data-points.

% We first sanitize the data by \textit{tokenizing} it. The process of tokenization breaks up code into pieces such as identifiers, keywords, constants, symbols and other elements. These pieces are called tokens. We use Python's {\tt javalang} module for tokenizing and parsing the code. We remove all data-points where either the $c_{old}$ or $c_{new}$ were not tokenizable by Python's {\tt javalang} module.
%% XXX can we say in a short sentence why code may not be tokenizable XXX
%This filters out 3,739 data-points. 

%\begin{comment}

%\end{comment}

Next, from the remaining $60,184$ data-points belonging to $16$ different templates, we remove data-points belonging to the three templates - `\textit{missing throws exception}', `\textit{delete throws exception}', `\textit{change modifier}' as $c_{old}$ and $c_{new}$ are missing for data-points of these templates in the dataset. We further remove data-points from template `\textit{change unary operator}' because the Path-Context Extractor provided by the authors of \cts{}~\cite{alon2018code2seq} was not able to capture $c_{old}$ and $c_{new}$ properly. Finally, we remove data-points of `\textit{change caller in function call}' template because they were already a part of the other templates.

%This leaves us with XXX data-points.
% I do not have the intermediate values

Though our design and implementation can randomly select $40$ path contexts from code, for the sake of this experiment, we filter out the data-points where the number of path-contexts were greater than $40$. The idea is to keep the input to the \edit{} model completely descriptive of the syntactic change, and therefore, give it every chance to outperform \lstm{} and \bow{}.

The filtering and pre-processing leaves us with $28,960$ data-points belonging to $11$ classes. Table \ref{tab:javadataset} provides an overview of these classes\footnote{\url{https://github.com/mast-group/mineSStuBs} contains the list of all the $16$ bug-templates, along with a brief description and the number of data-points for each bug-template.}. This number may seem like a relatively small set of data-points for deep-learning-based techniques. However, as we explain in Section \ref{sec:discussion}, code edit classification inherently suffers from a shortage of well-labeled data, a fact that previous work has also confirmed \cite{karampatsis2019often}. 

We divide the $28,960$ data-points into $26,322$ training samples and $2,638$ testing samples. We perform stratified sampling so that the proportion of each bug template or class in the training and test set are equal.

\subsubsection{Code Transformation Classification}
\label{ctclassification}
We created a dataset of edits generated by a set of C\# code transformers named {\em Rosylnator analyzers}~\cite{RoslynatorAnalyzers}, which analyze source code, list out parts which are not in compliance with a rule, and transform them if there is a code fix. For instance, analyzer $RCS1049$ transforms input source code \texttt{if (var1 == false) } to \texttt{if (!var1)}. Given $c_{old}$ and $c_{new}$, this task is to predict the the {\em Rosylnator analyzer} that was applied on $c_{old}$ to generate $c_{new}$.

Our dataset contains $12,784$ code edits generated by applying $10$ analyzers. Table \ref{tab:CSharpDataset} provides an overview of this dataset%(RCS1001, RCS1032, RCS1049, RCS1085, RCS1123, RCS1124, RCS1146, RCS1163, RCS1168 and RCS1220)
\footnote{The analysers are described in \url{https://github.com/JosefPihrt/Roslynator/tree/master/docs/analyzers}} over top $250$ C\# projects from GitHub based on their popularity. The edits are annotated with the Roslynator analyzer that has generated them.

We divide the $12,784$ obtained data-points into $11,617$ training samples and $1,167$ test samples. Here too, we perform stratified sampling so that the proportion of samples belonging to any Roslynator analyzer is the same in training and test set.

\begin{table*}[t]
 \centering
 \begin{tabular}{|c|c|c|}
 \hline
 Analyzer tag & Description & No of samples \\ [0.5ex] 
 \hline\hline
 RCS1001 & Add braces (when expression spans over multiple lines) & 443 \\ 
 \hline
 RCS1032 & Remove redundant parentheses & 516 \\
 \hline
 RCS1049 & Simplify boolean comparison & 574 \\
 \hline
 RCS1085 & Use auto-implemented property & 2163 \\
 \hline
 RCS1123 & Add parentheses according to operator precedence & 1428 \\
 \hline
 RCS1124 & Inline local variable & 1067 \\
 \hline
 RCS1146 & Use conditional access & 3368 \\
 \hline
 RCS1163 & Rename unused parameter to `\_' & 2053 \\
 \hline
 RCS1168 & Change parameter name to base name when they are not the same & 816 \\
 \hline
 RCS1220 & Use pattern matching instead of combination of 'is' operator and cast operator & 356 \\
 \hline
 \end{tabular}
 \vspace{0.1in}
 \caption{Descriptions of various analyzer tags, and the number of samples associated with the tags.}
\label{tab:CSharpDataset}
\end{table*}

%Similar to the Java dataset, this dataset too contains edits where the number of path-contexts in code\textsubscript{old} and code\textsubscript{new} are less than 40.

\section{Evaluation}
\label{sec:evaluation}

In this section, we compare the performance of the \cts{}-based model (\edit{}) with the baseline models described in Section \ref{sec:network} for the two code edit classification tasks described in Section~\ref{sec:exp}. 

To evaluate the performance of the models, we train the model with the optimal set of hyperparamaters obtained from hyperparamater tuning and use the average classification accuracy values of $3$ runs of $10$-fold cross-validation. Table \ref{tab:java_dataset_results} contains evaluation results for both bug-fix classification and the code transformation classification tasks.

The format of the table is as follows: the first column shows the classification model. The second and third column lists the average accuracy values for bug-fix classification with and without canonicalization (described later in this section) the last two columns lists the same for code transformation classification respectively. 

For both tasks, \lstm{} outperforms the other models in terms of classification accuracy (without canonicalization). Clearly, the \lstm{} model which considers code as a sequence of tokens improves the classification accuracy significantly over using \bow{}. This is because many code edits are position-sensitive and the sequence of tokens plays a significant role in classification. For instance, one example of `\textit{swap arguments}' class has the source code before edit ($c_{old}$) = \texttt{waitForJobExecutor(3000, 500)} and source code after edit ($c_{new}$) = \texttt{waitForJobExecutor(500, 3000)}. The only change in the code before and after the edit is the order of arguments in the method call. Such an order or sequence of tokens is well handled by \lstm{} model but \bow{} fails to capture this kind of change as the set of tokens for source code both before and after the edit remains the same.

% \usepackage{multirow}

% \begin{table}[t]
%  \centering
%  \begin{tabular}{|c|c|c|c|c|}
%  \hline
%  Model & Acc-Java & Acc-Java\textsubscript{can} & Acc-C\# & Acc-C\#\textsubscript{can}  \\ [0.5ex] 
%  \hline
%  tf-idf SVM (RBF) & 32.34\% & 58.81\% & 26.31\% & 26.31\% \\
%  \hline
%  tf-idf SVM (linear) & 85.30\% & 67.37\% & 85.30\% & 85.30\% \\
%  \hline
%  count SVM (RBF) & 32.34\% & 72.06\% & 34.24\% & 34.24\% \\
%  \hline
%  count SVM (linear) & 86.69\% & 76.08\% & 88.39\% & 88.39\%\\
%  \hline
%  \textbf{LSTM} & \textbf{94.47\%} & \textbf{99.21\%} & \textbf{92.55\%} & \textbf{92.77\%} \\
%  \hline
% % LSTM-attention model & 93.79\% & 99.17\% \\
%  %\hline
%  \cts{} & 93.17\% & 98.44\% & 92.28\% & 92.28\%\\ 
%  \hline
%  %\hline\hline
%  %code2seq-based model (canonicalized) & 98.44\% \\ 
%  %\hline
%  %LSTM model (canonicalized) & 99.21\% \\
%  %\hline
%  %LSTM-attention model (canonicalized) & 99.17\% \\
%  %\hline
%  \end{tabular}
%  \vspace{0.1in}
% \caption{Classification accuracy for the bug-fix classification and code transformation classification task}
% \label{tab:java_dataset_results}
% \end{table}

The \edit{} model, which captures the syntactic structure of code using path-contexts, does improve the classification accuracy (without canonicalization) over the \bow{} model but it does not outperform the \lstm{} model. At first, we found these results counter-intuitive as we expected to see an improvement in accuracy with the use of syntactic structure. As the difference between the accuracy values of both the models is not much, we perform statistical significance tests to confirm if the difference in the performance is statistically significant. For both \lstm{} and \edit{}, we first perform normality test over $30$ accuracy values, from $3$ runs of $10$-fold cross validation, using the D'Agostino-Pearson test \cite{d1973tests}, testing the null hypothesis that the accuracy values are normally distributed. We obtain a p-value of $0.51$ and $0.41$ for \lstm{} and \edit{} respectively, proving that they are indeed normally distributed. Since both the distributions are normal, we perform Student's t-test~\cite{Kalpic2011} on the $30$ accuracy values, for the \lstm{} and \edit{} models, the null hypothesis being that the distributions are identical. The p-value is $2.4*10^{-11}$ which clearly shows that the distributions are not identical, and the difference in the performance of the two models is statistically significant.

\begin{figure*}[!h]
\centering

\begin{subfigure}{0.5\textwidth}
\includegraphics[width=\textwidth]{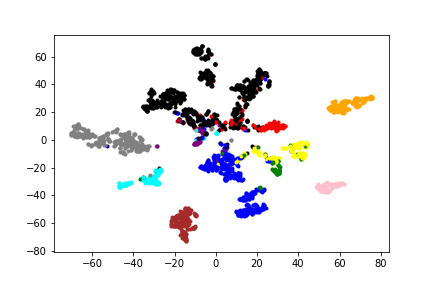}
\caption{t-SNE plot of \lstm{}}
\label{fig:tsne_LSTM_run1_fold1}
\end{subfigure}%
\begin{subfigure}{0.5\textwidth}
\includegraphics[width=\textwidth]{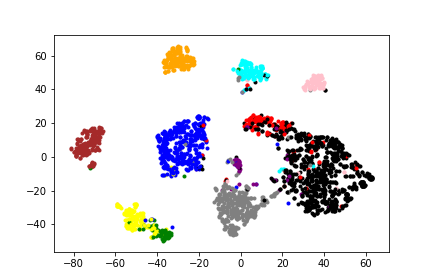}
\label{fig:tsne_AST_run1_fold1}
\caption{t-SNE plot of \edit{}}
\label{fig:tsne2}
\end{subfigure}

\begin{tabular}[l, columns = 3]{llllllll}
\\
\textcolor{purple}{$\blacksquare$} & \textit{swap arguments} &
\textcolor{pink}{$\blacksquare$} & \textit{overload method deleted args} &
\textcolor{green}{$\blacksquare$} & \textit{change operand} &
\textcolor{gray}{$\blacksquare$} & \textit{overload method more args} \\
\textcolor{blue}{$\blacksquare$} & \textit{change numeral} &
{$\blacksquare$} & \textit{different method same args} &
\textcolor{brown}{$\blacksquare$} & \textit{less specific if} &
\textcolor{cyan}{$\blacksquare$} & \textit{swap boolean literal} \\
\textcolor{orange}{$\blacksquare$} & \textit{more specific if} &
\textcolor{red}{$\blacksquare$} & \textit{change caller in function call} &
\textcolor{yellow}{$\blacksquare$} & \textit{change operator} &

\end{tabular}
%}

\caption{t-SNE plots of code edits from 11 classes of \textit{ManySStuBs4J dataset}   }
\label{fig:tSNE}
\end{figure*}

To further understand why \lstm{} does better than \edit{}, we examine, through visualizations, how well-separated and ``clean" the outputs from the two models are. As the data-points belong to $11$ different classes, we expect the code edit representations to form $11$ clusters in the intermediate latent space. So, we visualize the outputs from the layer before the \texttt{softmax} classification layer for both \edit{} and \lstm{}. We reduce these output vectors to $2$-dimensional vectors using t-SNE \cite{maaten2008visualizing}, and plot them, as shown in Figure \ref{fig:tSNE}. From the plots, we see that some classes of edits like `\textit{swap arguments}' (shown in light blue) and `\textit{overload method deleted args}' (shown in pink) are clearly separated from each other but there is significant overlap of the clusters for `\textit{change caller in function call}' (red) and `\textit{different method same args}' (black). We manually looked at some of the examples from these two classes and found that they are very similar to each other, which makes their separation more difficult from other classes. For instance, in `\textit{change caller in function call}' class, \texttt{var1.var2(param1,} \texttt{param2)} is changed to \texttt{var3.var2(param1, param2)}  where as in `\textit{different method same args}',\texttt{var1.var2(param1, param2)} is changed to \texttt{var1.var3(} \texttt{param1, param2)}.

Although both \edit{} and \lstm{} were not able to differentiate properly between many clusters, we also observed that \edit{} was not able to distinguish properly between some fairly different edits (for instance, classes `{\em different method same args}' (black) and `{\em overload method more args}' (grey)), which were well separated by \lstm{}.

% We initially conjectured that \edit{} was unable to distinguish between two similar edits, as the reason for this counter-intuitive observation. For example, the edits belonging to the classes \textit{CHANGE CALLER IN FUNCTION CALL} and \textit{DIFFERENT METHOD SAME ARGS} are similar to each other. To verify this, we visualize the outputs from the layer before the softmax layer, for both \edit{} and \lstm{}. We reduce these output vectors to 2-dimensional vectors, using t-SNE \cite{maaten2008visualizing}, and plot them. The plots for two different runs of \edit{} and \lstm{} are shown in Figure \ref{fig:tSNE}.

% Although it is true that \edit{} was not properly distinguishing between similar edits, we also observed that it was not able to distinguish properly between some fairly different edits (for example, \edit{} isn't able to properly segregate edits from classes `DIFFERENT METHOD SAME ARGS' and `OVERLOAD METHOD MORE ARGS'. This behaviour of \edit{} motivated us to further investigate for the reason.

To understand why \edit{} does not perform as well as \lstm{} for fairly different edits, we manually analyzed some of the data-points from the test set which were correctly classified by \lstm{} but incorrectly by \edit{}. One such case is shown in Figure \ref{fig:example}. The type of change (\textit{different method same args}) is the same in both the cases. Even the path\textsubscript{old} and path\textsubscript{new} are the same, the only difference is the left and right context tokens. The \lstm{} model classified both correctly. \edit{} classifies the first example correctly but fails to the classify the second example correctly. We found many similar examples in our manual analysis.
%We conjecture this is because of the worse class separation in \edit{} as compared to \lstm{}.

%These two datapoints were classified correctly by the LSTM model but \edit{} 
% and \edit{} have significantly higher accuracy values compared to \bow{} model.
% The LSTM model performs the best on both the Java and the C\# dataset. We performed statistical significance tests, and found that the comparison between all pairs of models are statistically significant. %We now describe our findings from this experiment.

%We analyzed samples from the test set, which were correctly classified by the LSTM model, but were wrongly classified by the code2seq-based model. A few examples are shown below.

% Consider the two examples shown in the two boxes. The type of change is the same in both the cases. path\textsubscript{old} and path\textsubscript{new} are also the same. The only difference is in the context tokens. The code2seq based model predicts the tag correctly in the first example, whereas, it predicts it incorrectly in the second example.

% \begin{figure*}
% \begin{center}
% \includegraphics[scale=0.6]{figures/code_encoder.pdf}
% \caption{Code Encoder}
% \label{fig:code_encoder}
% \end{center}
% \end{figure*}
% \begin{figure}[t]
%      \centering     
%      \includegraphics[width=\columnwidth]{figures/sysdesign.png}
%      \caption{\rex{} system design.}
%      \label{fig:design}
%  \end{figure}

\begin{figure*}[t]
\centering
\begin{subfigure}{0.9\columnwidth}
\begin{tcolorbox}[valign=top, halign=left, height = 5.7cm]
\textbf{c\textsubscript{old}} : \texttt{url.toDecodedString();}
\\ \textbf{c\textsubscript{new}} : \texttt{url.toString();}
\\ \textbf{left-context\textsubscript{old}} : url
\\ \textbf{left-context\textsubscript{new}} : url
\\ \textbf{right-context\textsubscript{old}} : toDecodedString
\\ \textbf{right-context\textsubscript{new}} : toString
\\ \textbf{path\textsubscript{old}} : NameExpression0, MethodCallExpression , NameExpression2
\\ \textbf{path\textsubscript{new}} : NameExpression0, MethodCallExpression, NameExpression2
\\ \textbf{tag\textsubscript{predicted}} : \textit{different method same args}
\\ \textbf{tag\textsubscript{actual}} : \textit{different method same args}
\end{tcolorbox}
\caption{Example correctly classified by \edit{}}
\label{fig:example1}
\end{subfigure}
%\hspace{0.1in}
\begin{subfigure}{0.9\columnwidth}
\begin{tcolorbox}[height = 5.7cm]
\textbf{c\textsubscript{old}} : \texttt{AtmResponse.create();}
\\ \textbf{c\textsubscript{new}} : \texttt{AtmResponse.newInstance();}
\\ \textbf{left-context\textsubscript{old}} : AtmResponse
\\ \textbf{left-context\textsubscript{new}} : AtmResponse
\\ \textbf{right-context\textsubscript{old}} : create
\\ \textbf{right-context\textsubscript{new}} : newInstance
\\ \textbf{path\textsubscript{old}}: NameExpression0, MethodCallExpression, NameExpression2
\\ \textbf{path\textsubscript{new}} : NameExpression0, MethodCallExpression, NameExpression2
\\ \textbf{tag\textsubscript{predicted}} : \textit{change caller in function call}
\\ \textbf{tag\textsubscript{actual}} : \textit{different method same args}
\end{tcolorbox}
\caption{Example incorrectly classified by \edit{}}
\label{fig:example2}
\end{subfigure}
\caption{Examples that are both correctly classified by \lstm{}, but only Example~\ref{fig:example1} is correctly classified by \edit{}.}
\label{fig:example}
\end{figure*}

Based on such manual analyses of test data that \lstm{} correctly classifies but \edit{} does not, we hypothesize that the \edit{} model relies heavily on the context tokens, and not as much on the path. Previous work~\cite{alon2019code2vec} too has made a similar observation where the authors did an ablation study to understand the contribution of each component of the path-context for the task of method name prediction and showed that the context tokens contributed significantly higher than the path of non-terminal nodes. The dependence of the model on the context tokens is good and, in fact, advantageous for method name prediction, which was the task targeted by both \cts{} and \ctv{}, because context tokens and method names have semantic relationships in a method. In contrast, semantics of context tokens serve no purpose in code edit classification, as the class of the code edit does not depend on individual tokens, but on the order of the tokens.

Consequently, to understand whether the token names were confounding our classification tasks, we repeat our experiments by dropping meaningful tokens in the path context and replacing them with canonical values. The canonicalization process is briefly explained as follows:
\begin{itemize}
\label{sec:canonicalize}

    \item Rename all identifiers with standard variable names like {\tt var1} and {\tt var2}. For example, \texttt{getConfig.getID()} is changed to \texttt{var1.var2()}.
    
    \item Replace all integer, float and string constants with standard integer, float and string constants For example, \texttt{getID(932,} \texttt{1044)} is changed to \texttt{var1(1,2)}; \texttt{rectangle.perimeter(2}\texttt{.345, 4.234)} is changed to \texttt{var1.var2(0.001, 0.002)}; \texttt{setName(}\texttt{"Alice", "Bob")} is changed to \texttt{var1("string1",} \texttt{"string2")}.
    %MAYBE WE CAN WRITE THIS MORE CLEARLY
\end{itemize}

We reevaluate the models with canonicalized context tokens for both the bug-fix classification  and code transformation classification task.  In the case of bug-fix classification, we see that the performance of \edit{} has improved significantly. But at the same time, the \lstm{} model has also improved. We confirmed that even with such canonical inputs, \lstm{} performs statistically better than the \edit{} model. Performing the Student' t-test, with the null hypothesis: accuracy values from \lstm{} and \edit{} have equal distributions, resulted in a p-value of $1.53*10^{-26}$. It again shows that the difference between the performance of the two models is statistically significant. With such canonicalization, the performance of the \bow{} model (linear classifier) reduces, which shows that \bow{} relies heavily on identifier names and not on the syntax of the code. For code transformation classification, though there is slight improvement in the performance of both \lstm{} and \edit{} after canonicalization, the improvement is not significant. This may be because of the larger code snippets in the code transformation dataset, when compared to the bug-fix dataset; the number of canonicalized variables in a code snippet are higher in the former case, defeating the purpose of canonicalization to some extent. In this case too, \lstm{} performs better than \edit{}.

% This suggests that the code2seq-based model relies heavily on the identifier tokens to classify the edit, instead of capturing the structural change. This has been observed in LSTM models as well, but not as frequently as in the code2seq-based model.

% To nullify the effect of identifier names, we reevaluate the models with canonicalized identifier names and constants for the Java dataset. The canonicalization process is briefly explained as follows:

We manually analyzed the mis-classified canonicalized data. For the canonicalized dataset, there were $30$ examples where the \lstm{} model classified correctly but the \edit{} failed to classify them correctly. But there were $10$ other examples which \edit{} classified correctly but \lstm{} model failed. By analyzing these examples, we learned that \edit{} captures some high-level syntactic representation of code which \lstm{} is not able to capture. Also, in Figure \ref{fig:tsne_LSTM_run1_fold1} for \lstm{} model, we see that the class `\textit{different method same args}' (black) is clustered into three sub-clusters, which is not the case for \edit{} (Figure \ref{fig:tsne2}). On manual analysis of some examples from each sub-cluster, we observed that they are grouped based on the number of arguments in a method. Methods having no arguments form one sub-cluster, methods with one argument form another sub-cluster and methods having more than one arguments form the third sub-cluster, even though they all belong to the same class. This sub-division within the cluster is not present for \edit{} model. This also supports that \edit{} is able to capture some high-level syntactic representation of code which \lstm{} fails to capture, but this representation is not sufficient to classify edit, specifically simpler and single-line edits. Single-line edits (specially SStuBs) seem to occur frequently~\cite{karampatsis2019often} compared to other types of code edits, and are easy to label, specifically in open-source code.

\begin{table}[t]
\begin{tabular}{|l|l|l|l|l|}
\hline
\multicolumn{1}{|c|}{\multirow{2}{*}{Model}} & \multicolumn{2}{c|}{Bug-fix} & \multicolumn{2}{c|}{Code Transformation} \\ \cline{2-5} 
\multicolumn{1}{|c|}{} & \multicolumn{1}{c|}{Accuracy} & \multicolumn{1}{c|}{\begin{tabular}[c]{@{}c@{}}Accuracy\\ (Canon.)\end{tabular}} & \multicolumn{1}{c|}{Accuracy} & \multicolumn{1}{c|}{\begin{tabular}[c]{@{}c@{}}Accuracy\\ (Canon.)\end{tabular}} \\ \hline
tf-idf SVM (RBF) & 32.34\% & 58.81\% & 26.31\% & 73.82\% \\ \hline
tf-idf SVM (linear) & 85.30\% & 67.37\% & 85.30\% & 69.49\% \\ \hline
count SVM (RBF) & 32.34\% & 72.06\% & 34.24\% & 76.31\% \\ \hline
count SVM (linear) & 86.69\% & 76.08\% & 88.39\% & 74.78\% \\ \hline
LSTM & 94.47\% & 99.21\% & 92.55\% & 92.77\% \\ \hline
code2seq & 93.17\% & 98.44\% & 92.28\% & 92.59\% \\ \hline
\end{tabular}
\caption{Classification accuracy for the bug-fix classification and code transformation classification task}
 \label{tab:java_dataset_results}
\end{table}
\section{Threats to Validity}
\label{sec:discussion}

While our findings do indicate that syntactic structure does not help code edit classification tasks, we recognize some threats to validity in this section. 

\subsection{Threats to Internal Validity}

In this section, we address threats to internal validity. 

\paragraph{\textbf{Lack of data.}} It is well-understood that present-day neural networks require large amounts of data to classify with high accuracy. The \lstm{} model has $269,147$ parameters, whereas the \edit{} model has $490,892$ parameters. \edit{} is more data-hungry, compared to \lstm{}. It is indeed possible that with significantly more data, \edit{} 's performance will improve significantly over what we report. 

However, collecting clean data for code edit classification is fundamentally difficult because it requires significant amounts of manual effort to confirm that labels for training data are indeed correct. It has been found~\cite{dyer2013boa} that commit messages and descriptions written by developers are very often not descriptive enough, and in many cases, they are completely meaningless. Also, developers tend to push multiple changes together in a commit, which makes it even more difficult to label smaller fixes within a commit. So fundamentally, we cannot use data from millions of commits of open source GitHub repositories. For our analysis, we use the ManySStuBs4J dataset \cite{karampatsis2019often}, which contains a particular set of classes of bug-fixes. These were collected from commits which have more descriptive commit messages, and their commit messages contain specific keywords like `error', `bug', `fix' etc. that help to decide if a commit is a bug-fix or not. However, this work shows that the average frequency of bug-fixes is about 1 per 1600-2500 lines of code, whereas the average length of a code snippet (method), as used by \texttt{code2seq}, is about 7 lines of code. Hence, narrowing in on bug-fixes reduces the data available by at least two orders of magnitude.

\paragraph{\textbf{Encoding specifics.}} We use the path-context as a fundamental unit of capturing syntactic structure. It is possible that a different representation of syntactic structure could improve the performance of \edit{}. However, our choice was driven by previous work which has successfully used the path-context construct~\cite{alon2019code2vec, alon2018code2seq}. It is also possible that a radically different model architecture will show different results. However, our choice of the model is done rigorously, by rigorously tuning the hyperparameters and choosing the best of 400 different models, and so we are confident that the design of the model will not significantly change our conclusions. 

\subsection{Threats to External Validity}

In this section, we address threats to external validity. While we have concentrated on two tasks across two languages, there may be other code edit classification tasks that could benefit from the use of syntactic structure. Also, our study is on relatively small code edits. It is possible that syntactic structure will help in classifying larger code edits. However, there is a tussle between larger code edits and accurately labelled data. Larger code edits make it even more difficult to collect accurately labelled data, since a summarization of the code edit would have to be more descriptive. Previous work~\cite{karampatsis2019often} have found that single-line edits occur relatively often (1 per 1600-2500 lines of code), making them potentially a promising dataset which can be used for various applications related to code edits.
%Hence we conjecture that our results will hold across multiple tasks that classify code edits, even if these edits are large. 

%In this section, we reason some of our findings from the experiments.

%There is a considerable difference in accuracies of the bag-of-words based classifier and the LSTM classifier, on both the datasets. It suggests that we can always leverage the sequence present in one-line code snippets, for classifying edits.

%Both the Java and the C\# datasets comprise of edits where the change has been made in a single line. The better performance of models that consider sequence (LSTM and LSTM with attention) compared to model that considers syntactic structure (code2seq based model) indicates that sequence matters more than the syntactic structure, when it comes to one-line code snippets.

%The code2seq model used for generating method names \cite{code2seq_paper} was trained on more than 15 million data-points. 
\section{Related Work}
\label{sec:relatedwork}

This paper takes inspiration from three categories of related work: using NLP techniques on code,  using syntactic structure to learn code embeddings and different ways of learning distributed representation of code edits.
In this section, we describe the related work in these areas.

\subsection{Using NLP Techniques On Code}
Inspired from the naturalness hypotheses and availability of enormous amount of code in the public domain, the use of natural language based embedding techniques on code has increased significantly where code is considered as a sequence of tokens.
\texttt{CODE-NN} by Iyer et al.~\cite{iyer-etal-2016-summarizing} presents an end-to-end neural attention model using LSTMs to generate summaries of C\# and SQL code. Their model outperforms the baseline models that use a tf-idf based approach. Li et al.~\cite{Cambronero_2019} compare various techniques including \texttt{CODE-NN} for neural code search and show that attention based weighting scheme on code embeddings outperforms more sophisticated techniques like \texttt{CODE-NN}. All of these approaches to learn embeddings of code tokens ignore the syntax of the code. They deal with well defined blocks of code (like functions or classes).

\subsection{Code Embeddings Using Syntactic Structure}
 As mentioned in Section~\ref{sec:introduction}, code has syntactic structure and long range correlations which make them different from natural language. Prior work has explored capturing such long range correlations by using data and control-flow graphs of code, abstract syntax trees~\cite{allamanis2017smartpaste, park2012using, nobre2016graph, zhang2019novel, allamanis2017learning}, etc.

Two of the most recent works in this space are \ctv{} \cite{alon2019code2vec} and \cts{} \cite{alon2018code2seq}. Both these techniques are used to represent entire code-snippets (methods) and are tested for method name prediction. In both these techniques, the code snippet is represented as a set of path-contexts extracted from its AST. An attentional mechanism is used over a set of randomly sampled path-contexts, while predicting or generating the method name. There are differences in the way embeddings are learnt for path-contexts. Path-context is considered as a single token in \ctv{}. Whereas, it is considered as a sequence of tokens in \cts{}. \cts{} has achieved the state of the art performance in code summarization (specifically method name prediction) and this inspired us to use this source code embedding technique to learn embeddings for code edits.

Gated graph neural networks (GGNNs) \cite{li2015gated} are another way to learn features for graph-structured inputs. Using these networks, ASTs are extended to graphs, by adding a variety of code dependencies as edges, to model code semantics~\cite{allamanis2017learning}. The tree based capsule network~\cite{jayasundara2019treecaps} is another such technique which captures both syntactic structure and dependency information in code, without the need of explicitly adding dependencies in the trees or splitting a big tree into smaller ones.  All of these techniques are used to represent entire code snippets, and not code edits.

\subsection{Learning Distributed Representation of Code Edits}

Most of the work related to code edits are targeted for specific applications like commit message generation, automatic program repair etc. The first works on commit message generation, by Loyola et al.~\cite{Loyola_2017} and Jiang et al.~\cite{Jiang_2017}, use attentional encoder-decoder architecture to generate commit messages from git diffs. Loyola et al.~\cite{Loyola_2017} use vanilla architecture where Jiang et al.~\cite{Jiang_2017} used a modified architecture with RNNs in the encoder. Liu et al.~\cite{10.1145/3238147.3238190} evaluate the performance of NMT-based techniques for commit message generation. They show that the performance of NMT-based techniques declines significantly when  automatically generated trivial commit messages were removed from the data. Liu et al.~\cite{10.1145/3238147.3238190} propose a simpler technique based on Nearest Neighbour algorithm (\texttt{NNGen}). In this case, the diff vectors are presented as a bag-of-words and cosine similarity between the diff vector of the input and the training data is used to find the nearest commit for the given input.

Closely related work to ours in this space is \texttt{commit2vec} \cite{lozoya2019commit2vec} where Lozoya et al. have compared AST based code representation learnt by \ctv{} \cite{alon2019code2vec} with other models based on LSTMs and bag-of-words, for binary classification of security related commits. They use transfer learning by using pre-trained embeddings from \ctv{} and another pretext task of predicting the priority of Jira tickets, both of which have large training data available. They observe that AST-based representation gives superior performance over other representations and pre-training the embeddings from a highly relevant pretext task further improves the results. They have tested it on a small dataset of 1950 commits. Also, they have used \ctv{}-based embeddings and the task is binary classification, In our case, we use \cts{}-based approach for multi-class classification problem on a much larger dataset from two different languages.

Another related work in this space is \texttt{DeepBugs} \cite{Pradel2017DeepLT}, a name based bug-detection technique. They model it as a binary classification problem and use semantic representations or embeddings of identifier names for classification. They also ignore the syntactic structure of the code while learning the embeddings.

Allamanis et al.\cite{yin2018learning} introduced the problem of learning distributed representation of edits. They propose a neural network based technique to combine the structure and semantics of code edits. They evaluate their model on the task of generating new code given old code and edit representation as input. Based on the results, they conclude that graph based edit encoder (which uses tree representation of code) often fails to capture simpler edits and it does not outperform the sequence based edit-encoder. Our observations from the results of code edit classification have similar conclusions; AST based representation of code does not help improve over the baseline approach of considering code as a sequence of tokens, i.e., \lstm{}

\section{Conclusion}
\label{sec:conclusion}
In this work, we introduced a code edit classification approach (\edit{}) that uses syntactic structure along with attention mechanism to learn distributed representation of code edits. We conducted experimental evaluations of this model on two tasks: bug-fix classification and code transformation classification and compared the performance of the model with baseline approaches that used \lstm{} and \bow{}. For both the tasks, we observe that the AST-based approach does not outperform the \lstm{} model. We observe that even though \edit{} captures
some high-level syntactic representation of code, which \lstm{} is not able to capture, this complex representation is not required in case of simpler and smaller edits. We believe that it is difficult to get large labelled datasets for code edits. While we have evaluated AST-based models on relatively small datasets of few thousand examples, further development is needed both in the area of getting labelled datasets and exploring these models for different tasks. We hope that our work inspires others to work in this interesting space.
\nocite{*}
\bibliography{paper}
\bibliographystyle{abbrv}

\end{document}

\newcommand{\edit}{{\tt edit2vec}}
\newcommand{\lstm}{{\tt LSTM}}
\newcommand{\bow}{{\tt Bag-of-words}}
\newcommand{\cts}{{\tt code2seq}}
\newcommand{\ctv}{{\tt code2vec}}

\title{Appendix}
\maketitle

\section{Dataset distribution tables}

\begin{table*}[t]
 \centering
 \begin{tabular}{|c|c|c|}
 \hline
 Analyzer tag & Description & No of samples \\ [0.5ex] 
 \hline\hline
 RCS1001 & Add braces (when expression spans over multiple lines) & 443 \\ 
 \hline
 RCS1032 & Remove redundant parentheses & 516 \\
 \hline
 RCS1049 & Simplify boolean comparison & 574 \\
 \hline
 RCS1085 & Use auto-implemented property & 2163 \\
 \hline
 RCS1123 & Add parentheses according to operator precedence & 1428 \\
 \hline
 RCS1124 & Inline local variable & 1067 \\
 \hline
 RCS1146 & Use conditional access & 3368 \\
 \hline
 RCS1163 & Rename unused parameter to `\_' & 2053 \\
 \hline
 RCS1168 & Change parameter name to base name when they are not the same & 816 \\
 \hline
 RCS1220 & Use pattern matching instead of combination of `is' operator and cast operator & 356 \\
 \hline
 \end{tabular}
 \vspace{0.1in}
 \caption{Descriptions of various analyzer tags, and the number of samples associated with the tags.}
\label{tab:1}
\end{table*}

\begin{table*}[t]
 \centering
 \begin{tabular}{|p{3cm}|p{8.5cm}|p{2cm}|}

% \begin{tabular}{|c|c|c|}
 \hline
 Bug-template & Description & No of samples \\ [0.5ex] 
 \hline\hline
 CHANGE CALLER IN FUNCTION CALL
 & Checks whether in a function call expression the caller object for it was replaced with another one.
 & 1488 \\ 
 \hline
 CHANGE NUMERAL
 &  Checks whether a numeric literal was replaced with another one & 4779 \\
 \hline
 CHANGE OPERAND
 & Checks whether one of the operands in a binary operation was wrong. & 741 \\
 \hline
 CHANGE OPERATOR & Checks whether a binary operator was accidentally replaced with another one of the same type & 1711 \\
 \hline
 DIFFERENT METHOD SAME ARGS
 & Checks whether the wrong function was called. & 9383 \\
 \hline
 LESS SPECIFIC IF
 & Checks whether an extra condition which either this or the original one needs to hold ($\∥$ operand) was added in an if statement’s condition. & 2095 \\
 \hline
 MORE SPECIFIC IF
 & Checks whether an extra condition ($\&\&$ operand) was added in an if statement’s condition & 1836 \\
 \hline
 OVERLOAD METHOD DELETED ARGS
 &  Checks whether an overloaded version of the function with less arguments was called & 1040 \\
 \hline
 OVERLOAD METHOD MORE ARGS
 & Checks whether an overloaded version of the function with more arguments was called & 3820 \\
 \hline
 SWAP ARGUMENTS
 & Checks whether a function was called with two of its arguments swapped & 536 \\
 \hline
SWAP BOOLEAN LITERAL
 &  Checks whether a Boolean literal was replaced & 1531 \\
 \hline
 \end{tabular}
 \vspace{0.1in}
 \caption{Descriptions of various SStUB templates, and the number of samples associated with the template.}
\label{tab:javadataset}
\end{table*}